\documentclass[10pt,twocolumn,letterpaper]{article}

\usepackage[pagenumbers]{cvpr} 

\usepackage{graphicx}
\usepackage{amsmath}
\usepackage{amssymb}
\usepackage{booktabs}
\usepackage{fancyvrb}

\usepackage{graphicx}
\usepackage{amsmath}
\usepackage{amssymb}
\usepackage{booktabs}
\usepackage[export]{adjustbox}
\usepackage{tikz}
\usetikzlibrary{arrows.meta,
                calc, chains,
                quotes,
                positioning,
                shapes.geometric}
\usepackage{placeins}
\usepackage{graphbox}
\usepackage{breqn}
\usepackage{multirow}
 \usepackage{mathtools}

\usepackage[pagebackref,breaklinks,colorlinks]{hyperref}
\usepackage[inline]{enumitem}

\usepackage[capitalize]{cleveref}
\crefname{section}{Sec.}{Secs.}
\Crefname{section}{Section}{Sections}
\Crefname{table}{Table}{Tables}
\crefname{table}{Tab.}{Tabs.}

\def\cvprPaperID{2} 
\def\confName{CVPR}
\def\confYear{2023}

\tikzstyle{startstop} = [rectangle, rounded corners, 
minimum width=3cm, 
minimum height=1cm,
text centered, 
draw=black, 
fill=red!30]

\tikzstyle{io} = [trapezium, 
trapezium stretches=true, 
trapezium left angle=70, 
trapezium right angle=110, 
minimum width=3cm, 
minimum height=1cm, text centered, 
draw=black, fill=blue!30]

\tikzstyle{process} = [rectangle, 
minimum width=3cm, 
minimum height=1cm, 
text centered, 
text width=3cm, 
draw=black, 
fill=orange!30]

\tikzstyle{decision} = [diamond, 
minimum width=3cm, 
minimum height=1cm, 
text centered, 
draw=black, 
fill=green!30]
\tikzstyle{arrow} = [thick,->,>=stealth]
\newcommand{\imwidth}{0.14\textwidth}
\newcommand{\newtexttoimagesamples}{
\begin{figure*}[htbp]
\centering

	\setlength{\tabcolsep}{1pt}
	\begin{tabular}{ccccccc}
	\toprule
	\multicolumn{7}{c}{\small{Fashion Content Synthesis}} \\
	\midrule

    \shortstack{\tiny{\emph{`Indian Fashion'}}} & \shortstack{\tiny{\emph{`African Fashion'}}} & \shortstack{\tiny{\emph{`Italian Fashion'}}} & \shortstack{\tiny{\emph{`Middle East Fashion'}}} & \shortstack{\tiny{\emph{`Icelander Fashion'}}}  & \shortstack{\tiny{\emph{`Balkan Fashion'}}} & \shortstack{\tiny{\emph{`Japanese Fashion'}}} \\

	\midrule
	\includegraphics[align=c,width=\imwidth]{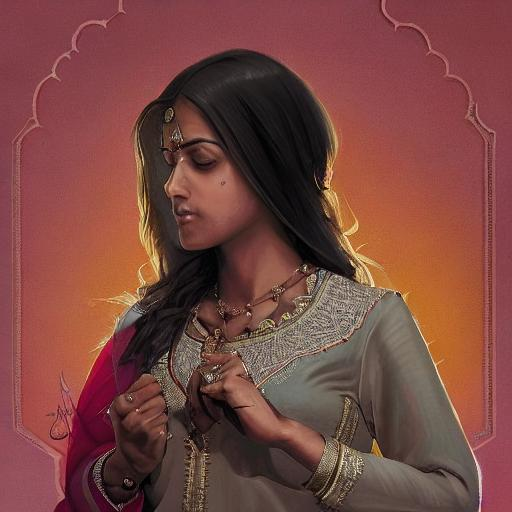} &
	\includegraphics[align=c,width=\imwidth]{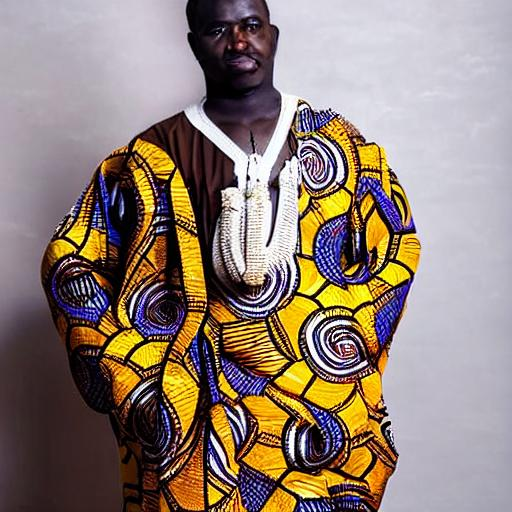} &
	\includegraphics[align=c,width=\imwidth]{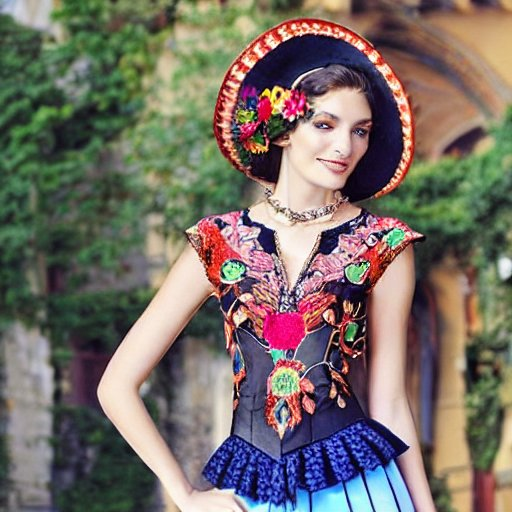} &
	\includegraphics[align=c,width=\imwidth]{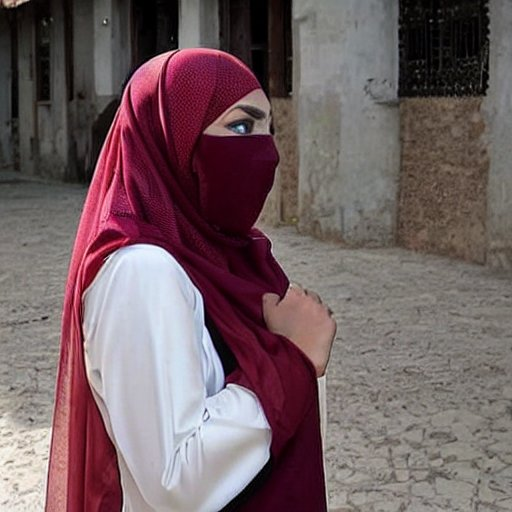} &
	\includegraphics[align=c,width=\imwidth]{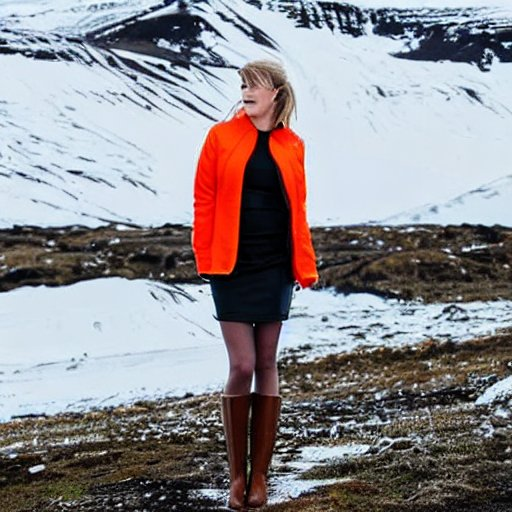} &
	\includegraphics[align=c,width=\imwidth]{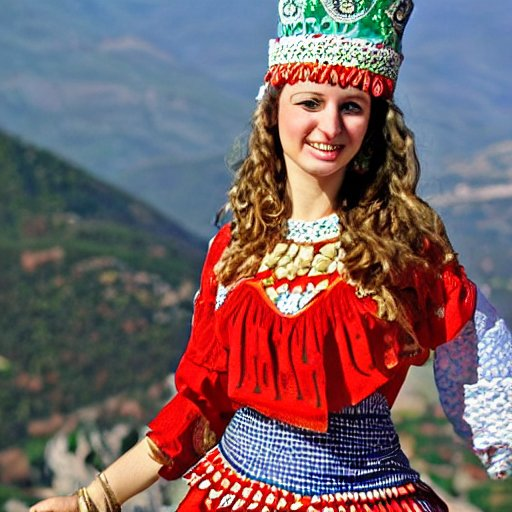} &
	\includegraphics[align=c,width=\imwidth]{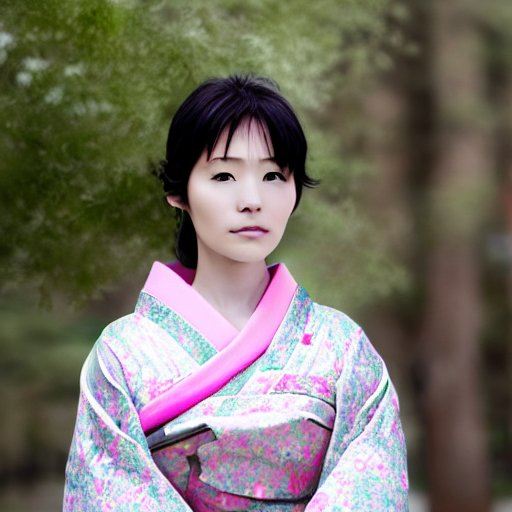} \\
	
	\includegraphics[align=c,width=\imwidth]{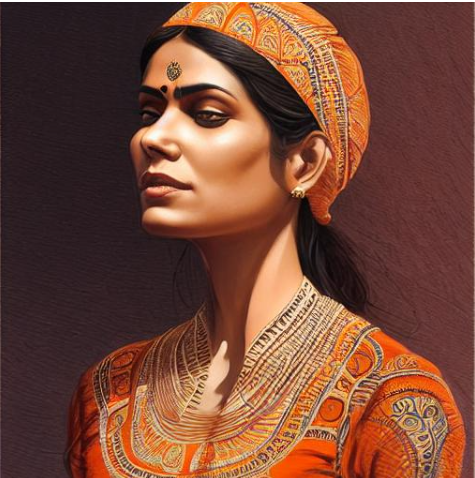} &
	\includegraphics[align=c,width=\imwidth]{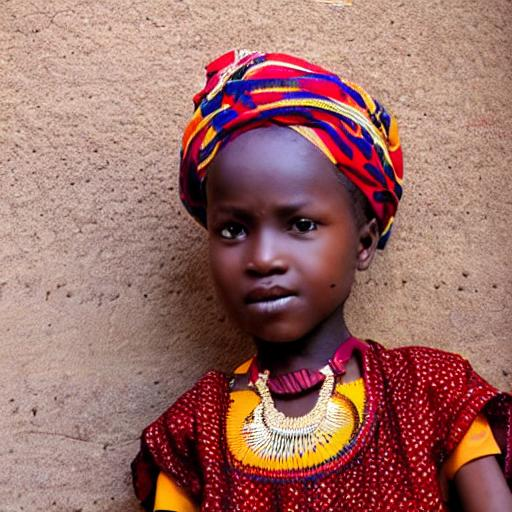} &
	\includegraphics[align=c,width=\imwidth]{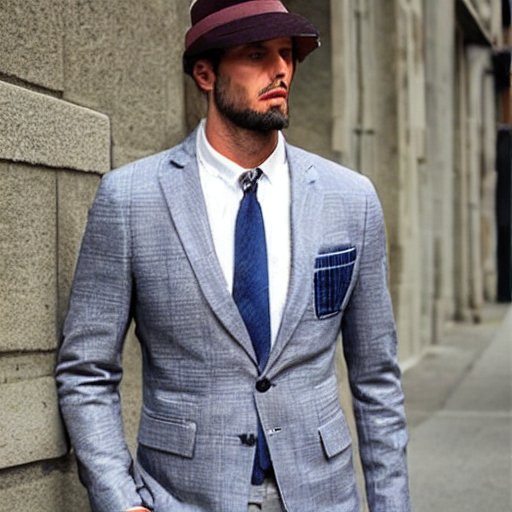} &
	\includegraphics[align=c,width=\imwidth]{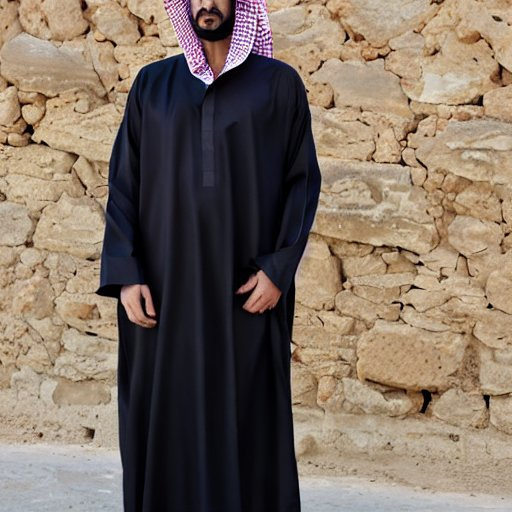} &
	\includegraphics[align=c,width=\imwidth]{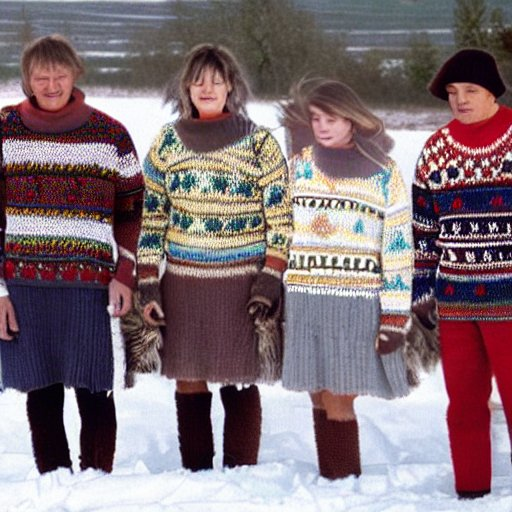} &
	\includegraphics[align=c,width=\imwidth]{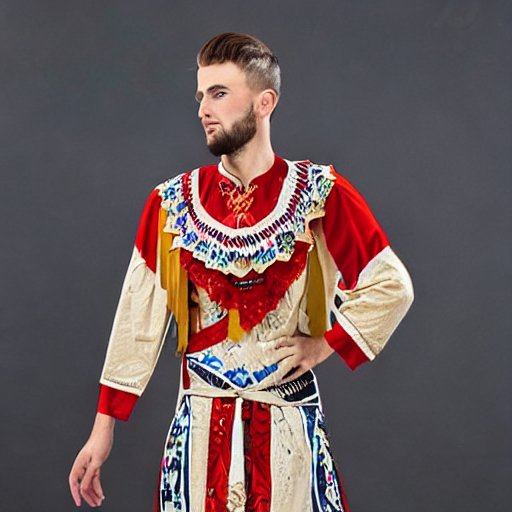} &
	\includegraphics[align=c,width=\imwidth]{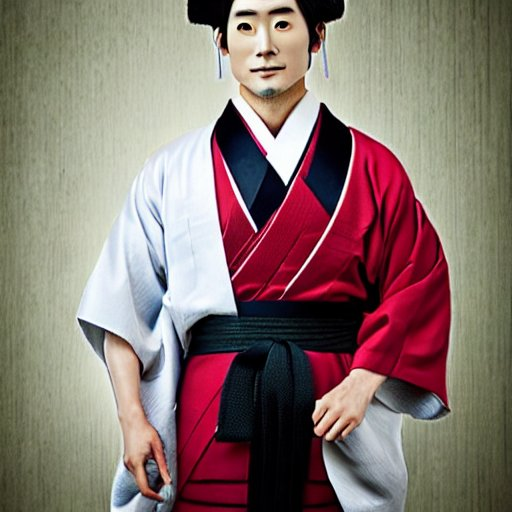} \\

	\bottomrule
	\end{tabular}
\caption{\label{fig:text2img_samples} Samples for user-defined text prompts from our model for text-to-image synthesis, \emph{LDM}, which was trained on the LAION~\cite{schuhmann2021laion} database and fine-tuned on DeepFashion~\cite{liuLQWTcvpr16DeepFashion}. Samples generated with 200 DDIM steps and $\eta = 1.0$. We use unconditional guidance \cite{ho2021classifierfree} with $s=10.0$.}
\end{figure*}
}

\begin{document}

\title{Interactive Fashion Content Generation Using LLMs and Latent Diffusion Models}


\author{Krishna Sri Ipsit Mantri\\
{\tt\small ipsit.iitb@gmail.com}
\and
Nevasini Sasikumar\\
PES University\\
Bangalore, India\\
{\tt\small nevasini24@gmail.com}
}
\maketitle

\begin{abstract}

    Fashionable image generation aims to synthesize images of diverse fashion prevalent around the globe, helping fashion designers in real-time visualization by giving them a basic customized structure of how a specific design preference would look in real life and what further improvements can be made for enhanced customer satisfaction. Moreover, users can alone interact and generate fashionable images by just giving a few simple prompts. Recently, diffusion models have gained popularity as generative models owing to their flexibility and generation of realistic images from Gaussian noise. Latent diffusion models are a type of generative model that use diffusion processes to model the generation of complex data, such as images, audio, or text. They are called "latent" because they learn a hidden representation, or latent variable, of the data that captures its underlying structure. We propose a method exploiting the equivalence between diffusion models and energy-based models (EBMs) and suggesting ways to compose multiple probability distributions. We describe a pipeline on how our method can be used specifically for new fashionable outfit generation and virtual try-on using LLM-guided text-to-image generation. Our results indicate that using an LLM to refine the prompts to the latent diffusion model assists in generating globally creative and culturally diversified fashion styles and reducing bias.
   
\end{abstract}

\section{Introduction}
\label{sec:intro}
Virtual try-on technology (also known as VTON) allows users to see how clothing and accessories would look on them without actually trying them on. This technology is especially useful for designers in interactive fashion content generation because it allows them to create more engaging and personalized content for their customers. By using VTON technology, designers can create interactive fashion content that allows users to visualize how different outfits and accessories would look on them. This helps customers to make more informed decisions when shopping online and can also increase their confidence in making purchases. It can also help designers to identify potential design issues before a garment goes into production. By allowing users to see how a garment looks and fits in a virtual environment, designers can make adjustments to the design and fit before the garment is produced, saving time and money. Overall, virtual try-on technology can be a valuable tool for designers in interactive fashion content generation. This technology helps in enhanced customer engagement, and improved design decisions and is a valuable tool for designers in interactive fashion content generation.


Existing 3D model-based VTON approaches produce photorealistic VTON results in 3D, but modeling and scanning for 3D is both expensive and time-consuming~\cite {minar2020cloth, bhatnagar2019multi}. With the advent of deep learning, 2D image-based approaches have become popular as the 2D data is easier to collect and maintain, and 2D requires less computing than 3D~\cite{song2020sp,wang2018toward}. But these suffer from variations in clothing and human style. This is where 3D model-based approaches excel. 3D models can easily reconstruct complex human poses and fabric deformation and illumination effects. However, VTON requires combining 3D human pose reconstruction and fabric reconstruction, which is a difficult task~\cite{mir2020learning,li2017learning,joo2018total}.

Immense progress has been made in generative models across various domains~\cite{brown2020language,brock2018large,ho2022classifier}. These models serve as excellent priors for downstream generation tasks like question-answering~\cite{brown2020language}, image-generation~\cite{https://doi.org/10.48550/arxiv.1812.04948}, and text-to-image generation~\cite{saharia2022photorealistic} and many more. But this has resulted in models with an increasingly large number of parameters and require huge datasets to train and finetune. Diffusion models~\cite{sohl2015deep,song2019generative,ho2020denoising} have become recently popular for generative modeling because of their sample quality, traceability, and flexibility. The sampling process can also be guided post-training of these models with~\cite{sohl2015deep,dhariwal2021diffusion} or without using classifiers~\cite{ho2022classifier}.

Generating high-quality images from text descriptions is a challenging task. It requires a deep understanding of the underlying meaning of the text and the ability to generate an image consistent with that meaning. To address this problem in this work, we propose
the use of an energy-based parameterization for latent diffusion
models~\cite{du2023reduce}, where the unnormalized density of each reverse
diffusion distribution is explicitly modeled. We further propose to use a pre-trained LLM to refine the text prompts that are being given to the latent diffusion model. We 
demonstrate the effectiveness of our approach in high-resolution VTON and creative fashion generation. 
\section{Related Work}
In this section, we review relevant prior work and discuss existing research on $(i)$ generative adversarial networks, $(ii)$ text--conditioned image generation and editing, $(iii)$ GAN inversion techniques, and $(iv)$ the use of computer vision in fashion. The goal of the section is to provide the necessary background for our work. More comprehensive coverage of these topics can be found in some of the recent surveys, e.g.,~\cite{gan_inversion_survey, wu2017survey, cheng2021fashion}.  

\subsection{Generative Adversarial Networks}
	Generative Adversarial Networks (GANs)~\cite{gan} have in recent years become the \textit{de-facto} method for  unconditional image synthesis, allowing convincing high-resolution image synthesis and reasonable training times with consumer-grade hardware. DCGAN~\cite{dcgan} introduced convolutional GANs and provided architectural pointers to achieve successful GAN convergence. ProGAN~\cite{progan} was the first GAN model that achieved megapixel-sized images thanks to a progressive learning scheme.
    StyleGAN~\cite{stylegan1} introduced a non-linear mapping of the latent space and an alternative generator design inspired by the style transfer literature. 
 StyleGAN2~\cite{stylegan2} further adjusted the generator architecture to remove the frequent droplet artifacts and regularized the training with path-length regularization. StyleGAN2-ADA~\cite{stylegan2-ada} proposed several augmentation techniques to enable learning a high-quality GAN with limited training data. StyleGAN3~\cite{stylegan3} presented a continuous interpretation of the generator signals to prevent the dependence of the generated image on the absolute pixel coordinates, in turn enabling more natural latent code interpolations.

Next to the architectural advances, considerable progress has also been made in the field of GAN regularization and training. Various techniques of stabilizing the GAN convergence by modifying the training objective were proposed in~\cite{wgan, wgan-gp, lsgan}. Other successful methods include weight normalization~\cite{miyato2018spectral} and various regularization approaches~\cite{mescheder2018training}. 

\subsection{Text--Conditioned Image Generation and Editing}

Text--conditioned image generation models are focused on generating realistic images that match the semantics of the provided text descriptions. Conversely, corresponding \textit{editing techniques} try to realistically manipulate images in a way that preserves the image characteristics irrelevant to the text description. Thus, text--conditioned image editing aims to alter only the semantic content that is expressed in the text description while preserving all other parts of the data. 

\textbf{Image Generation.} The seminal work of Reed \textit{et al.}~\cite{reed2016generative} proposed a text--conditioned GAN model by feeding the text information to both the generator and discriminator of the GAN design. StackGAN~\cite{stackgan} and StackGAN++~\cite{stackgan++} proposed stacked generators, where the resolution of the generated images increased progressively with each generator in the stack. AttnGAN~\cite{attngan} proposed an attention mechanism to attend to relevant words, on which to  condition the image--generation process. MirrorGAN~\cite{mirrorgan} proposed a cyclic GAN architecture that regularized the generated image by enforcing correct (re)descriptions. 

While these early models discussed above already provided for competitive performance, more recent text--conditioned generative models are several orders of magnitude larger in size and are trained on several orders of magnitude larger datasets. DALL-E~\cite{dalle}, for example, trained a $12$-billion parameter autoregressive transformer on $250$ million of image--text pairs, and considerably outperformed previous models in the considered (zero--shot) evaluation experiments. The model was further improved in DALL-E 2~\cite{dalle2} through the use of diffusion techniques. Imagen~\cite{imagen} proposed to increase the size of the text encoder to achieve better results in terms of image fidelity and image--text alignment.

\textbf{Image Editing.} An early approach to text--conditioned image editing methods was described by Dong \textit{et al.} in~\cite{dong2017semantic}. Here, the authors proposed a conditional encoder-decoder GAN model and reported impressive results on two diverse datasets. Nam \textit{et al.}~\cite{nam2018text} proposed so-called word-subset local discriminators that enabled fine-grained image editing. ManiGAN~\cite{li2020manigan} proposed a different strategy for merging image and text representations while adding a detail correction module for enhanced image quality. 

Another notable group of methods performs image editing by first converting a given image into the latent
code of some pre-trained GAN, in a process known as GAN inversion, then performing various latent code manipulations to achieve the desired edits. InterFaceGAN~\cite{interfacegan}, for example, identified directions in the latent space of StyleGAN2 that corresponded to specific semantic changes (given by binary attribute labels) in the corresponding output image. Image2StyleGAN~\cite{image2stylegan, image2stylegan++} performed several face image edits using GAN inversion, and StyleCLIP~\cite{styleclip} proposed different methods for text-guided image editing, where the general idea is based on combining
the generative capabilities of StyleGAN with the image--text matching capabilities of the CLIP model~\cite{clip}. TediGAN~\cite{tedigan} introduced a control mechanism based on style mixing in StyleGAN to achieve the desired semantics in face images driven by text descriptions. 

\subsection{GAN Inversion} 

As can be seen from the literature review presented above,  a considerable amount of existing editing techniques deals with the process of GAN inversion to retrieve the image's latent code for editing. How to conduct the GAN inversion is a key consideration with these techniques that has a significant impact on the final editing capabilities. 
Richardson \textit{et al.}~\cite{psp}, for example, proposed an encoder, called pSp, for projecting images into the StyleGAN latent space and demonstrated its feasibility through several image-to-image translation tasks. E4e~\cite{e4e} adjusted the pSp model so that the latent codes follow a similar distribution as the original StyleGAN latent codes and performed image edits with several latent code manipulation techniques.
ReStyle~\cite{restyle} presented an iterative procedure to obtain the latent code, while HyperStyle~\cite{hyperstyle} adjusted the StyleGAN generator weights on a per-sample basis to achieve better image reconstructions.
Additional GAN inversion methods can be found in the recent survey~\cite{gan_inversion_survey}.

\subsection{Diffusion Models}
Recently, \textbf{Diffusion Probabilistic Models} (DM)~\cite{pmlr-v37-sohl-dickstein15}, have achieved state-of-the-art results in density estimation \cite{kingma2021on} as well as in sample quality \cite{NEURIPS2021_49ad23d1}. The generative power of these models stems from a natural fit to the inductive biases of image-like data when their underlying neural backbone is implemented as a UNet~\cite{ronneberger2015unet, ho2020denoising, song2021scorebased, NEURIPS2021_49ad23d1}.
The best synthesis quality is usually achieved when a reweighted objective \cite{ho2020denoising}
is used for training. In this case, the DM corresponds to a lossy compressor and allow to trade image quality for compression capabilities.
Evaluating and optimizing these models in pixel space, however, has the downside of low inference speed and very high training costs.
While the former can be partially addressed by advanced sampling strategies~\cite{song2020denoising, sanroman2021noise, kong2021on} and hierarchical approaches~\cite{ho2021cascaded, vahdat2021score}, training on high-resolution image data always requires to calculate expensive gradients.
%
%
%
 
\subsection{Computer Vision in Fashion} 
A considerable cross-section of fashion-related computer-vision research focuses on Virtual Try-On technology, where, given an image of a person and some target garment, the goal is to realistically fit the garment while preserving the original person's pose and appearance.
VITON~\cite{viton}, CP-VTON~\cite{cp-vton} proposed to warp the target clothing by conditioning the warping procedure on a coarse human shape and pose map before blending it with the input image. VTNFP~\cite{vtnfp} and C-VTON~\cite{fele2022c} adopted semantic segmentation models to guide the synthesis. Yang \textit{et al.}~\cite{yang2020towards} constrained the fashion item warping and introduced adaptive content generation and preservation constraints. MG-VTON~\cite{mg-vton} proposed a network that enabled multi-pose try-on. VITON-HD described a
Virtual Try-On method for higher resolution image generation. DCTON~\cite{cyclefashion} utilized cycle--consistency for the editing procedure~\cite{cycle}, and WUTON~\cite{wuton} and PF-AFN~\cite{pf-afn} proposed a teacher-student setup that removed the need for intermediate auxiliary  models (e.g., for parsing, pose-estimation) during the editing step.

Work has also been performed to enable text conditioning for editing or generating target images. FashionGAN~\cite{prada}, for example, edited images conditioned on text inputs, segmentation masks, and several image-specific categorical attributes using encoder-decoder GANs. Fashion-Gen~\cite{fashiongen} introduced a dataset of the image--text pairs and experimented with unconditional and text--conditioned image generation. Recently, Text2Human~\cite{text2human} proposed generating human images based on a description of clothes' shape and texture. With this approach, the text--encoding method mapped the input text into a number of closed sets of categories, which limits the language expressiveness of the input text.

Unlike the methods presented above, our work mainly focuses on removing bias and helping in a uniform and fair generation of diversified fashionable images. Preventing bias in image synthesis is an important ethical consideration in the development of artificial intelligence. It is very crucial to train a model on a diversified and representative dataset of different demographics. This can help prevent the model from learning and reproducing biased patterns. Hence we use the DeepFasion dataset~\cite{iccv2017fashiongan} to fine-tune our StableDiffusion model, which consists of data samples belonging to different countries, cultures, and communities. This helps in preserving cultural diversity and traditions specific to every country and, further, helps local designers to build interactively upon the generated images from VTON. It also helps the common man to generate highly personalized fashionable images specific to a country from just a few simple prompts.

\section{Background}
\label{sec:background}
\subsection{Diffusion Models}
Diffusion models are a class of latent variable-based generative models that differ from other generative models in the following ways
\begin{enumerate*}
    \item The dimensionality of the latent variable is the same as that of the original data
    \item The procedure in which the latent variables are learned is fixed via a Markov Chain
    \item Exact modeling of original data distribution and exact sampling
\end{enumerate*}
Consider $x_0 \sim q(x_0)$ sampled from a data distribution $q$. A diffusion model aims to model $q$ by creating a Markov Chain and adding Gaussian noise in every state as per the following (this is called the \textit{forward process})
\begin{align}
    q(x_t | x_{t-1}) &= \mathcal{N}(x_t; \sqrt{1 - \beta_t}x_{t-1}, \beta_t \boldsymbol{I}) \label{eqn:markov}\\
    q(x_{1:T} | x_0) &= \prod\limits_{t=1}^T q(x_t | x_{t-1})
\end{align}
Here $\big\{\beta_t \in \left(0, 1\right]\big\}_{t=1}^T$ is a schedule of pre-defined constants which resemble the step size of the diffusion process and $T$ denotes the number of diffusion steps. For large enough $T$, $x_T \sim \mathcal{N}(0, \bold{I})$. The reverse process is in which we start from a random noise sample $x_T$ and try to reconstruct the original data point $x_0$. We do this by parametrizing the reverse distribution $q(x_{t-1} | x_t)$ using a model $p_\theta(x_{t-1}| x_t)$ with parameters $\theta$. For a small enough $\beta_t$~\cite{https://doi.org/10.48550/arxiv.1503.03585}, $p_\theta(x_{t-1}| x_t) \sim \mathcal{N}(x_{t-1}; \mu_\theta(x_t,t), \tilde{\beta}_t \bold{I})$ where
\begin{align}
    \mu_\theta(x_t, t) &= \frac{1}{\sqrt{\alpha_t}} \bigg(x_t - \frac{\beta_t}{\sqrt{1 - \bar{\alpha}_t}\, \epsilon_\theta(x_t, t)}\bigg)
\end{align}
with $\epsilon_\theta(x_t, t)$ denoting our neural network and $\alpha_t,  \bar{\alpha}_t, \tilde{\beta_t}$ being functions of $\big\{\beta_t \in \left(0, 1\right]\big\}_{t=1}^T$ alone.

Using \cref{eqn:markov}, it is easy to see that 
\begin{align}
    q(x_t | x_0) &= \mathcal{N}(x_0; \sqrt{\bar{\alpha}_t}x_0, (1 - \bar{\alpha}_t)\,\bold{I})
\end{align}
So for $x_t = \sqrt{\bar{\alpha}_t}x_0 + \epsilon \, \sqrt{(1 - \bar{\alpha}_t)}$ where $\epsilon \sim \mathcal{N}(0, \bold{I})$, we train the diffusion model by minimizing the objective $\mathcal{L} = \sum\limits_{t=1}^T L_t$:
\begin{align}
    L_t &= \mathbb{E}_{\mathcal{N}(\epsilon; 0, \bold{I})}\big[\|\epsilon - \epsilon_\theta(x_t, t)\|_2^2\big]
    \label{eqn:difft}
\end{align}
After training, we can generate data samples from the original distribution $q(x_0)$ starting from random Gaussian noise $x_T$ and using $p_\theta(x_{t-1}|x_t)$.
\begin{figure}[ht]
    \centering
    \begin{tikzpicture}[node distance=2cm, auto, every node/.style={ auto, font=\footnotesize, anchor=center}]
\centering
\node (in1) [io] {Initial Prompt from Designer};
\node (pro1) [process, below of=in1] {MagicPrompt2 (LLM)};

\node (pro2) [process, below of=pro1] {StableDiffusion (LDM)};

\node (out1) [io, below of=pro2] {Generated Fashion Image};

\draw [arrow] (in1) -- (pro1);
\draw [arrow] (pro1) -- (pro2);
\draw [arrow] (pro2) -- (out1);

\end{tikzpicture}
    \caption{Proposed Method}
    \label{fig:my_label}
\end{figure}
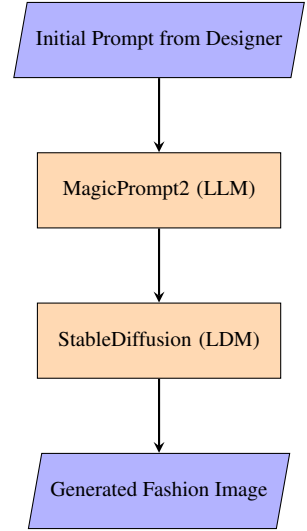
\subsection{Energy Based Models}
Energy Based Models (EBMs) are used for density estimation. Given a dataset $x \sim q(x)$, EBMs parameterize the distribution as $p_\theta(x) = \frac{-\exp(f_\theta(x))}{Z(\theta)}$ where $Z(\theta)$ is the partition function and $f_\theta(x) \in \mathbb{R}$ is any neural network. EBMs are flexible as it is not necessary to explicitly model the partition function $Z(\theta)$ and can be trained using alternate equivalent optimization objectives instead of maximizing the log-likelihood as for other generative models.

Denoising score matching is a popular method for training EBMs. The score function of a probability distribution is defined as 
\begin{align}
    s_\theta(x) &= \nabla_x \log p_\theta(x)
\end{align}. In the context of EBMs, $s_\theta(x) = -\nabla_x f_\theta(x)$. In denoising score matching, we minimize the Fisher divergence between the Gaussian noise corrupted data $q_\sigma(x) = \int q(y)\,\mathcal{N}(x;y,\bold{I})\:dy$ and the model
\begin{align}
    \mathcal{J} &= \mathbb{E}_{q(x)\,\mathcal{N}(\epsilon;0,\bold{I})}\bigg[\bigg\|\frac{\epsilon}{\sigma} + \nabla_x f_\theta(x + \sigma\,\epsilon)\bigg\|_2^2\bigg]
\end{align}
\subsection{Sampling From EBM}
To sample from the learned EBM, we rely on Markov Chain Monte Carlo (MCMC) based methods. In this method, we generate samples by simulating a Markov chain starting from a point $x_0$ sampled from a known prior distribution. We transition between states using an ergodic and invariant transition function $T(x_t | x_{t-1})$.

A most commonly used variant of MCMC for EBMs is the Unadjusted Langevin Dynamics Algorithm (ULA)~\cite{roberts1996exponential}, which is given by
\begin{align}
    x_t &\xleftarrow{} x_{t-1} + \delta \nabla_x f_\theta(x_{t-1}) + \sqrt{2\delta}\,z_t\\
    z_t &\sim \mathcal{N}(0, \bold{I})
\end{align}
i.e., $T(x_t | x_{t-1}) = \mathcal{N}(x_t; x_{t-1} + \delta \nabla_x f_\theta(x_{t-1}), 2\delta\,\bold{I})$. ULA will eventually generate samples from actual distribution only when the step size $\delta$ is small. To speed up the sampling process, i.e., to enable larger step sizes, we incorporate Metropolis correction and use Metropolis-Adjusted-Langevin-Algorithm~\cite{grenander1994representations} given by
\begin{align}
    x_t &\xleftarrow{} \lambda \, \hat{x} + (1 - \lambda)\, x_{t-1}\\
    \hat{x} &\sim \mathcal{N}(x_{t-1} + \delta \nabla_x f_\theta(x_{t-1}), 2\delta\,\bold{I})\\
    \lambda &= \min \bigg(1, \frac{e^{f_{\theta}(\hat{x})}}{e^{f_\theta(x_{t-1})}}\,\frac{T(x_{t-1}|\hat{x})}{T(\hat{x}|x_{t-1})}\bigg)
\end{align}
\newtexttoimagesamples
\subsection{Diffusion Models and EBMs}
Consider optimizing an EBM using Denoising Score Matching (ignoring the constant factors)
\begin{align}
    J_t &= \frac{1}{\sigma^2_t}\mathbb{E}_{\mathcal{N}(\epsilon; 0, \sigma^2_t\bold{I})}\big[\|\sigma_ts_\theta(x_0+ \sigma_t \epsilon) + \epsilon\|_2^2\big]
    \label{eqn:dsm}
\end{align}
If we compare~\cref{eqn:dsm} with~\cref{eqn:difft}, we have $\epsilon_\theta(x_t, t) = -\sqrt{(1 - \bar{\alpha}_t)}\,s_\theta(x_0+ \sigma_t \epsilon) = -\sqrt{(1 - \bar{\alpha}_t)}\,\nabla_x \log q_\sigma(x)$. Hence we can conclude that $\nabla_x f_\theta(x, t) = -\frac{\epsilon_{\theta}(x_t, t)}{\sqrt{(1 - \bar{\alpha}_t)}}$ where $f_\theta(x, t)(x, t)$ is our noise conditional EBM.
\subsection{Conditional Generation}
We can introduce a trained predictive model $p_\theta(y|x;t)$, which is conditioned on time predicting the distribution of a label $y$. We can then use this trained model to guide (with guidance parameter $\lambda$) the sample generation by exploiting the Bayes rule as follows:
\begin{align}
    \nabla_x \log p_\theta(x|y;t) &= \nabla_x \log p(x;t) + \lambda \nabla_x \log p_\theta(y|x;t)
\end{align}

\section{Proposed Method}

We propose to use conditional text to image generation using the StableDiffusion Model~\cite{rombach2021highresolution} (referred to as LDM) to generate culturally diverse new fashion styles. To give the input text prompt, we use the GPT-2 based MagicPrompt model~\cite{MagicPrompt}. The flowchart can be seen in \cref{fig:my_label}. More details about each module follow:
\subsection{Prompt Generation using MagicPrompt}
\begin{figure}[htbp]
    \centering
    \includegraphics[scale=0.35]{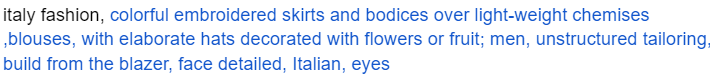}
    \caption{Designer prompt, \color{blue}refined prompt from LLM}
    \label{fig:stage1}
\end{figure}

We use a pre-trained large language model (LLM) to refine the input prompts to the conditional image-generating diffusion model. The motivation behind using an LLM is that large models like LLMs and diffusion models can learn even without gradient flow, i.e., implicit learning only based on prompting. The designer will first give a text input regarding fashion to MagicPrompt, and it will generate a more descriptive prompt that will help in making the designer's idea about fashion concrete. 
An example of such prompting can be seen in ~\cref{fig:stage1}

\subsection{Conditional Image Generation using Text}
In this stage, we give the generated text from MagicPrompt to the pre-trained StableDiffusion model. The StableDiffusion model is pre-trained and further fine-tuned on the DeepFasion dataset~\cite{liu2016deepfashion} and uses Fashion Synthesis Benchmark~\cite{iccv2017fashiongan}; it includes 78,979 images selected from the In-shop Clothes Benchmark. Each image is associated with several sentences as captions and a segmentation map. The LDM has following stages:
\begin{enumerate}
    \item \textbf{Variational Autoencoder (VAE)}: 
    

    The encoder of VAE is used to transform the image into low-dimensional space, which will be used as input to the UNet. The decoder transforms the latent image back into the original dimension. Encoder is used during the forward diffusion process to generate latent representations, whereas the decoder is used during inference in the reverse diffusion process to re-transform latent representations into original images. Hence during inference, we only need the decoder.
    \item \textbf{UNet}: Both the encoder and decoder parts of UNet are ResNet based. The main role of the UNet is denoising the image, i.e., the encoder transforms the image into a lower dimension, and the decoder upsamples the image back to its original shape, possibly leading to a less noisy version. To enable conditioning, we add cross-attention layers to the encoder and decoder between the ResNet blocks. Finally, the output of the UNet predicts the noise residual and is used to compute the predicted image representation.
    \item \textbf{Text Encoder}: We use a trained text encoder~\cite{https://doi.org/10.48550/arxiv.2103.00020} to encode the input text prompt and pass it as input to the UNet. 
    
\end{enumerate}

\section{Experiments \& Results}
We experiment using two different settings, first, we use the LDM trained on the LAION dataset to directly generate fashion images, and second, we fine-tune the pre-trained LDM on the DeepFashion dataset. We use the default training settings as in \cite{rombach2021highresolution} to fine-tune the LDM. As seen in \cref{tab:tab1}, the fine-tuned model generates better images with lower FID scores and higher Inception Scores. 
\begin{table}[htbp]
\centering
\begin{footnotesize}
\begin{adjustbox}{max width=\linewidth}
\footnotesize
\begin{tabular}{lcc}
\toprule
\textbf{Method} & FID$\downarrow$ & IS$\uparrow$ \\
\midrule
%
%
%
%
%
%
LLM + LDM (w/o fine-tuning) & 10.55 & 28.75\tiny$\pm\text{0.46}$ \\
\emph{LLM + LDM} (ours) & \textbf{7.80} & \textbf{31.64\tiny$\pm\text{2.33}$} \\
\bottomrule
\end{tabular}
\end{adjustbox}
\end{footnotesize}
\caption{Quantitative Results for DeepFashion Dataset~\cite{iccv2017fashiongan}}
\label{tab:tab1}
\end{table}

\begin{figure}[ht]
    \centering
    \includegraphics[scale=0.3]{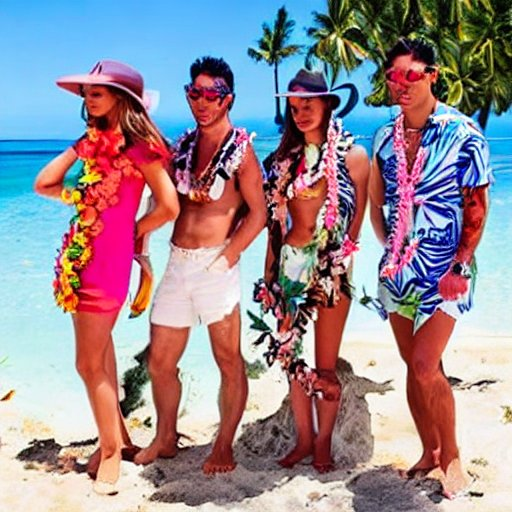}
    \caption{Generated Image with Distorted Body Parts}
    \label{fig:my_distorbed}
\end{figure}

From the images in \cref{fig:text2img_samples}, it can be seen that refining prompts using a pre-trained LLM helps in generating high-resolution, photo-realistic, and creative fashion styles which were otherwise not possible. We can further conclude that the diffusion not only generates 2D VTON samples but it generates images with folds in the garment and visual effects along with proper human body features and style pose. Our approach not just generates the garments but also generates VTON samples. One shortcoming of this work is that the results in \cref{fig:text2img_samples} don't represent the facial fashion styles well enough. Some of the generated fashion images have body features (\cref{fig:my_distorbed}), especially faces, and their facial organs are distorted. We plan to tackle this weakness in our future work. 

\section{Conclusion}

In this paper, we use the diffusion model pre-trained on LAION-5B multi-modal dataset and fine-tune it on the high-resolution fashion image DeepFashion dataset. Our focus is that the model should be very representative and diversified in terms of generalizing fashion across the globe. This can help to prevent any bias and preserve the cultural and traditional design styles of every country and promote cultural awareness. Further, this can support local artisans and designers who specialize in traditional clothing styles, can help preserve their craft and keep traditional styles alive and try making them more real-time and interactive and build more creative designs styles keeping the generated image from VTON as a base. 

\section{Future Work}

We plan to extend VTON to generate high-quality, fashionable 3D images again, working on making it bias-free. 3D image synthesis involves generating images that have depth, volume, and other physical properties associated with objects in a 3D space. These images can be generated from a 3D model of an object or scene and then render into a 2D image. The process involves modeling objects in a 3D space, adding textures and lighting, and then rendering the final fashionable image.

{\small
\bibliographystyle{ieee_fullname}
\bibliography{main}
}

\end{document}